\documentclass[letterpaper, 10 pt, conference]{ieeeconf}  

\usepackage{cite}
\usepackage{graphicx}
\usepackage{caption}
\usepackage{subcaption}
\usepackage{amsmath}
\usepackage{amsfonts}
\usepackage{amssymb}
\usepackage{booktabs}
\usepackage{tabularx}
\usepackage{array}
\usepackage{multirow}
\usepackage{hyperref}
\usepackage{dashrule}
\usepackage{tikz}

\newcommand{\ie}{\textit{i.e.}}

\IEEEoverridecommandlockouts                              

\title{\LARGE \bf
BiCo-Fusion: Bidirectional Complementary LiDAR-Camera Fusion for Semantic- and Spatial-Aware 3D Object Detection
}

\author{Yang Song$^{1}$ and Lin Wang$^{2,*}$
\thanks{*Corresponding author}
\thanks{$^{1}$Yang Song is with the AI Thrust, The Hong Kong University of Science and Technology (Guangzhou), Guangdong 511458, China.
        {\tt\small yangsong1105@gmail.com}}%
\thanks{$^{2}$Lin Wang is with the School of Electrical and Electronic Engineering (EEE), Nanyang Technological University (NTU), Singapore, Email:
        {\tt\small alwang.ntu@gmail.com}}
}

\begin{document}

\maketitle
\pagestyle{empty}
\thispagestyle{empty}

\begin{abstract}

3D object detection is an important task that has been widely applied in autonomous driving. To perform this task, a new trend is to fuse multi-modal inputs, \ie, LiDAR and camera. Under such a trend, recent methods fuse these two modalities by unifying them in the same 3D space.
However, during direct fusion in a unified space, the drawbacks of both modalities (LiDAR features struggle with detailed semantic information and the camera lacks accurate 3D spatial information) are also preserved, diluting semantic and spatial awareness of the final
unified representation.
To address the issue, this letter proposes a novel bidirectional complementary LiDAR-camera fusion framework, called \textbf{BiCo-Fusion} that can achieve robust semantic- and spatial-aware 3D object detection. 
The key insight is to fuse LiDAR and camera features in a bidirectional complementary way to enhance the semantic awareness of the LiDAR and the 3D spatial awareness of the camera. The enhanced features from both modalities are then adaptively fused to build a semantic- and spatial-aware unified representation.
Specifically, we introduce Pre-Fusion consisting of a Voxel Enhancement Module (VEM) to enhance the semantic awareness of voxel features from 2D camera features and Image Enhancement Module (IEM) to enhance the 3D spatial awareness of camera features from 3D voxel features. We then introduce Unified Fusion (U-Fusion) to adaptively fuse the enhanced features from the last stage to build a unified representation. 
Extensive experiments demonstrate the superiority of our BiCo-Fusion against the prior arts. Project page: \url{https://t-ys.github.io/BiCo-Fusion/.}

\end{abstract}


\vspace{-6pt}
\section{INTRODUCTION}

3D object detection \cite{bevfusion_icra23, transfusion, uvtr} is a critical and challenging task that aims to localize and classify objects in the 3D space, which has been widely applied in applications such as robotics and autonomous driving. Early works often resorted to one single sensor as inputs, such as LiDARs or RGB cameras. While approaches based on these two sensors have yielded significant results, they yet need to address and overcome several inherent limitations.
LiDAR-based methods \cite{voxelnet,pointpillars,centerpoint,focalsparse,fastpoinrrcnn,voxelsettrans,3dssd,lsk3dnet,pillarbased,voxelrcnn,std,second,pointgnn,pointrcnn,frustum} have difficulty classifying small objects due to poor performance on detailed semantic information, while camera-based methods \cite{lss,m3d-rpn,fcos3d,bevdet,bevdepth,bevformer,detr3d,petr} perform even worse, struggling with localization due to the lack of accurate spatial and depth information, especially when the objects are occluded~\cite{sparsefusion}.

\begin{figure*}[htbp]
    \centering
    \begin{minipage}[b]{0.28\textwidth}
        \centering
        \includegraphics[width=\textwidth]{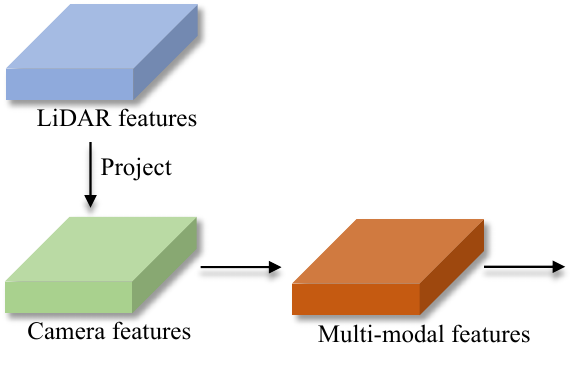}
        \caption*{(a) 2D-plane fusion }
    \end{minipage}
    \hspace{0.001\textwidth}
    \begin{tikzpicture}
    \draw[dashed] (0,-0.3) -- (0,3.5);
    \end{tikzpicture}
    \hspace{0.001\textwidth}
    \begin{minipage}[b]{0.28\textwidth}
        \centering
        \includegraphics[width=\textwidth]{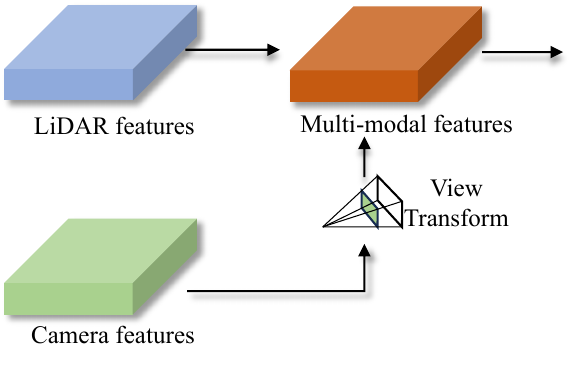}
        \caption*{(b) 3D-space fusion}
    \end{minipage}
    \hspace{0.001\textwidth}
    \begin{tikzpicture}
    \draw[dashed] (0,-0.3) -- (0,3.5);
    \end{tikzpicture}
    \hspace{0.001\textwidth}
    \begin{minipage}[b]{0.35\textwidth}
        \centering
        \includegraphics[width=\textwidth]{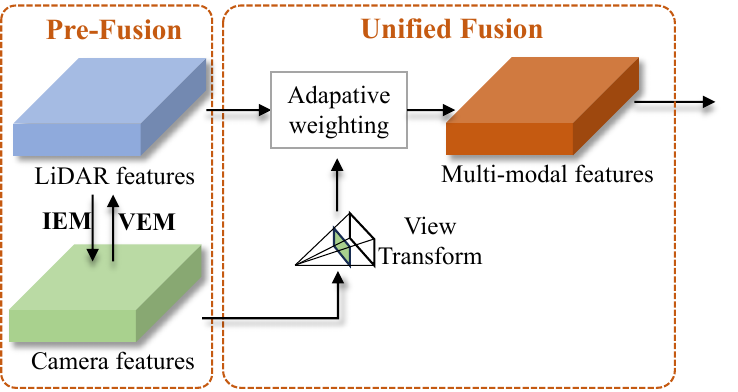}
        \caption*{(c) Our BiCo-Fusion}
    \end{minipage}
    \vspace{-2pt}
    \caption{ \textbf{Comparison of our framework with previous LiDAR-camera fusion methods.} (a) 2D-plane fusion projects point clouds or proposals generated from them to the image plane and retrieves 2D features for fusion. (b) 3D-space fusion lifts camera features by view transformation to a unified space with LiDAR features, where the two modalities are fused. (c) Our BiCo-Fusion proposes Pre-Fusion with a bidirectional complementary strategy to compensate for the
drawbacks of both modalities, including VEM and IEM which enhance semantic awareness for LiDAR and 3D spatial awareness for the camera, respectively. During Unified Fusion, the enhanced features are unified in a 3D space by view transformation for further adaptive-weighting-based fusion.}
    \label{framework_brief}
    \vspace{-18pt}
\end{figure*}

A new trend to handle these problems is to fuse two modalities, aiming to utilize more complementary information from both sparse point clouds and dense camera data. In the context of this trend, existing methods to deal with the fusion of two heterogeneous modalities can be divided into two main categories: \textit{i}) 2D-plane fusion (Fig.~\ref{framework_brief}(a)) and \textit{ii}) 3D-space fusion (Fig. \ref{framework_brief}(b)). Early 2D-plane fusion methods \cite{pointpainting,MV3D,EPNet,deepfusion,pointaugmenting,multitask,transfusion} perform LiDAR-camera fusion by projecting point clouds or proposals generated from them to the image plane and retrieving the corresponding 2D camera features, which are used to decorate the 3D representations in order to compensate for the semantic ability. However, the camera features are much denser, such a projection only considers fusing sparse point clouds or proposals with part of them, which wastes semantically rich 2D features~\cite{msmdfusion}.

While recent 3D-space fusion methods unify two modalities in the same 3D space~\cite{bevfusion_nips22,bevfusion_icra23,uvtr}, either in voxel \cite{uvtr} or in bird’s eye view (BEV) space \cite{bevfusion_icra23}, and fuse them to obtain a unified representation. This significantly solves the problem of 2D-plane fusion as all dense camera features can be transformed into 3D space to participate in the fusion with LiDAR.
However, it cannot be ignored that both modalities have drawbacks. Specifically, LiDAR features struggle with detailed semantic information while the camera falls short of accurate 3D spatial information. These drawbacks are also inevitably preserved during direct fusion in a unified space. As a result, this dilutes semantic and spatial awareness of the final unified representation, which are modality-specific strengths inherited from the camera and LiDAR, respectively.


In this letter, we propose \textbf{BiCo-Fusion}, a novel multi-modal 3D object detection framework to tackle the above problem. BiCo-Fusion employs a bidirectional complementary strategy to fuse LiDAR and camera features to enhance the semantic awareness of the LiDAR and the 3D spatial awareness of the camera. It then adaptively fuses the enhanced features from both modalities to construct a semantic- and spatial-aware unified representation, as shown in Fig. \ref{framework_brief}(c). Firstly, the extracted features from LiDAR and the camera are input into the initial stage of fusion, referred to as \textit{Pre-Fusion} (Sec. \ref{prefusion}). Here, two modules work together in a bidirectional complementary way to play crucial roles: the Voxel Enhancement Module (VEM) projects non-empty voxels onto the image plane and extracts the surrounding camera features to enhance their semantic awareness, which is optimized based on our distance-prior weighting scheme.
The Image Enhancement Module (IEM) fuses the real depth from the LiDAR with camera features to enhance their 3D spatial awareness. Through Pre-Fusion, the drawbacks of both modalities undergo our specific enhancements, preparing more comprehensive LiDAR and camera features for the subsequent process.
Next, we unify the enhanced camera features to the same 3D space as LiDAR by view transformation \cite{lss}. Finally, we introduce an adaptive weighting scheme to perform \textit{Unified Fusion (U-Fusion)}, which dynamically fuses the enhanced features from both modalities to build a unified representation (Sec. \ref{unifiedfusion}). Overall, these components work seamlessly to guarantee a robust semantic- and spatial-aware 3D representation for the detection.

Extensive experiments on the nuScenes \cite{nuscenes} benchmark demonstrate that our BiCo-Fusion approach achieves state-of-the-art (SOTA) performance, with 72.4\% mAP and 74.5\% NDS on the test set.
In summary, our contributions are three-fold: (\textbf{I}) We propose a novel BiCo-Fusion framework that enables a semantic- and spatial-aware unified representation for 3D object detection; (\textbf{II}) We propose Pre-Fusion works in a bidirectional complementary way, including VEM for enhancing the semantic awareness of LiDAR voxels and IEM for enhancing the 3D spatial awareness of camera features, and Unified Fusion for adaptively unifying representations in the 3D space; (\textbf{III}) We achieve the SOTA 3D detection performance on the challenging nuScenes dataset.

\begin{figure*}[thpb]
      \centering
      \includegraphics[width=\textwidth]{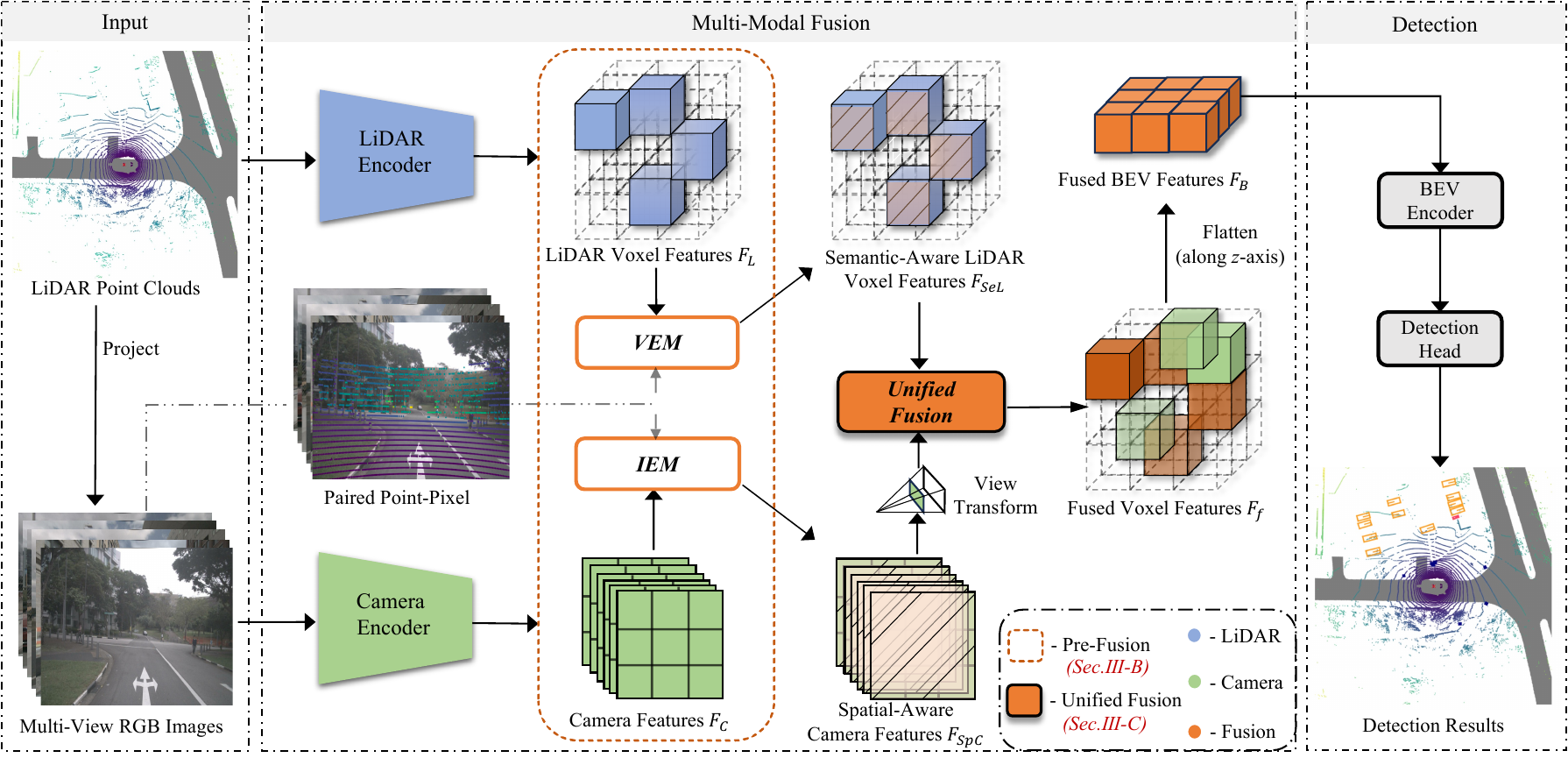}
      \caption{\textbf{Overview of our BiCo-Fusion framework}. BiCo-Fusion first extracts features from LiDAR and camera data using modality-specific encoders. 
      In Pre-Fusion, the LiDAR voxel features are enhanced with the camera semantics by our VEM, and the camera features are enhanced to be spatial-aware with the IEM. 
      These enhanced features are then fused in an adaptive way during the Unified Fusion stage. Finally, the fused features are flattened to get the BEV features, which are fed to the head for final detection.}
      \label{framework}
      \vspace{-18pt}
\end{figure*}

\vspace{-8pt}
\section{RELATED WORK}
\vspace{-2pt}
\noindent\textbf{LiDAR-based 3D Object Detection.} LiDAR point clouds are inherently suitable for 3D object detection, in that they can provide accurate 3D spatial information. However, the point cloud data is sparse and unordered, which results in point cloud data that cannot be directly fed into a CNN network for feature extraction. PointNet \cite{pointnet} is the first backbone capable of extracting point cloud features, inspiring a great deal of subsequent works \cite{pointrcnn,fastpoinrrcnn,pointgnn}. Later, to obtain more regular point cloud data, researchers choose to rasterize point cloud data into discrete grid representation, such as voxels \cite{voxelnet,voxelrcnn,second} and pillars \cite{pointpillars}, enabling direct feature extraction with convolutional networks. However, they cannot break the limitation of using a single modality and are prone to false detections due to missing semantics in special cases, such as small and distant objects.

\noindent\textbf{LiDAR-Camera Fusion-based 3D Object Detection.} This has become a new trend, aiming to maximize the use of both worlds. Early 2D-plane fusion methods (Fig.~\ref{framework_brief}(a))~\cite{pointpainting,pointaugmenting,MV3D,transfusion,deepfusion} typically use a projection strategy to retrieve corresponding camera features for decorating point clouds or proposals that are generated from them. PointPainting \cite{pointpainting} is a pioneering work that performs fusion at the point level, which directly projects raw point clouds into the image plane for fusion. While MV3D \cite{MV3D} chooses to fuse camera features with the projected proposals, denoted as proposal-level methods \cite{transfusion}. However, regardless of point-level or proposal-level methods, they only associate the sparse LiDAR point clouds or proposals with part of the dense camera features, wasting semantically rich information in them.

Recently, 3D-space fusion methods (Fig. \ref{framework_brief}(b)) \cite{msmdfusion,bevfusion_icra23,uvtr,cmt,unitr,bevfusion_nips22} try to construct a unified representation for both modalities by unifying them in a shared 3D space for fusion. This paradigm naturally mitigates the problem of 2D-plane fusion as all dense camera features can be transformed into the shared space to engage in the fusion. UVTR \cite{uvtr} fuses the LiDAR and camera features in a unified voxel space. While BEVFusion \cite{bevfusion_icra23} generates a unified representation by unifying both modalities in the birds' eye view (BEV) space to improve the 
computational efficiency, which provides a solid pipeline for recent works~\cite{msmdfusion,cmt,unitr,sparsefusion,deepinteraction}.
However, these methods ignore that during direct fusion in a unified space, the drawbacks of both modalities (LiDAR features struggle with detailed semantic information and the camera lacks accurate 3D spatial information) are also preserved. It dilutes semantic and spatial awareness of the final unified representation.

\textit{To this end, our BiCo-Fusion (Fig.~\ref{framework_brief}(c)) proposes Pre-Fusion with a bidirectional complementary strategy, including Voxel Enhancement Module (VEM) to enhance the semantic awareness of LiDAR features and Image Enhancement Module (IEM) to enhance the 3D spatial awareness of camera features. This helps to ensure the comprehensiveness of the
two modalities at the semantic and spatial levels, beneficial for building up a unified representation in our Unified Fusion (U-Fusion) stage.}


\section{METHODOLOGY}
\subsection{Framework Overview}
\label{framework_overview}
\vspace{-3pt}
As shown in Fig. \ref{framework}, given the raw LiDAR and RGB camera data as inputs, we first extract features from specific encoders. Following BEVFusion \cite{bevfusion_icra23}, we use VoxelNet \cite{voxelnet} as the LiDAR encoder and Swin Transformer \cite{swintransformer} as the camera encoder, respectively. 

\begin{figure}[t!]
      \centering
      \includegraphics[width=.98\columnwidth]{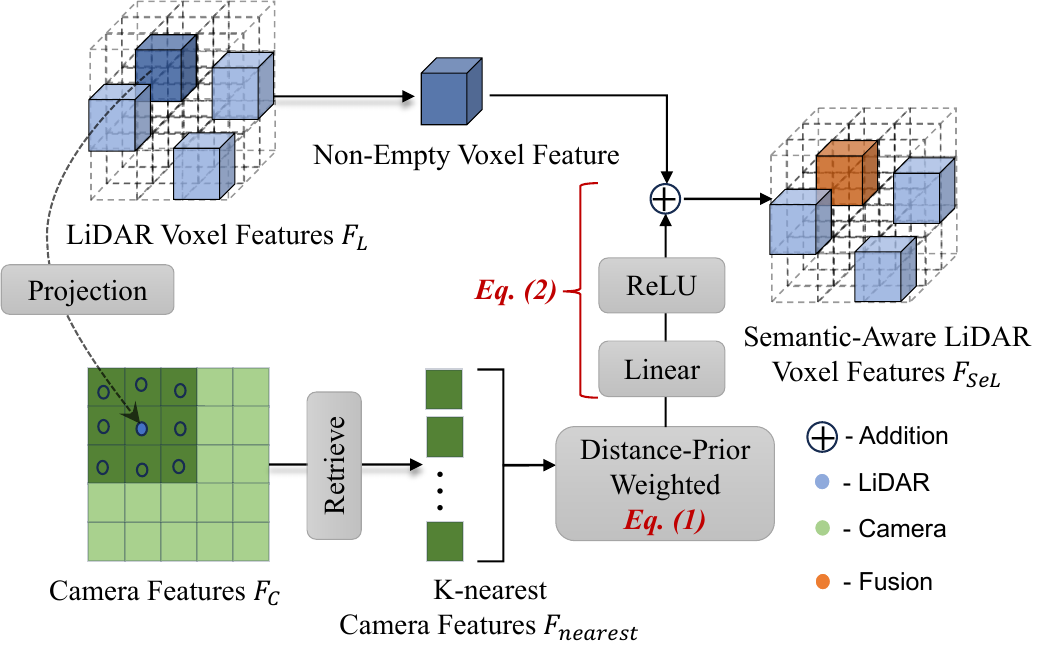}
      \vspace{-5pt}
      \caption{\textbf{Voxel Enhancement Module (VEM).} VEM first projects voxel to the image plane to retrieve the \(K\) nearest camera features. Then, the camera features are weighted in a distance-prior way, which then be inputted to a Linear layer for achieving a learnable process. Finally, the resulted camera features are fused with the original LiDAR voxel features by addition.}
      \label{vem} 
      \vspace{-25pt}
\end{figure}

Then, the extracted LiDAR voxel features $F_L$ and camera features $F_C$ are interacted with and fused using our designed method. Our multi-modal fusion space contains two fusions, in the Pre-Fusion phase, our Voxel Enhancement Module (VEM) and Image Enhancement Module (IEM) perform enhancements for both modalities in a bidirectional complementary way to improve the semantic capability of voxels and the spatial capability of images, which effectively compensates for their drawbacks, and also prepares more comprehensive features for the next step. Afterward, in the Unified Fusion phase (U-Fusion), the enhanced spatial-aware camera features $F_{SpC}$ then be unified into the 3D voxel space by view transformation \cite{lss,bevfusion_icra23,uvtr}, which are further fused with the semantic-aware LiDAR voxel features $F_{SeL}$ adaptively to build a unified representation $F_f$.

Following the workflow of voxel-based approaches \cite{voxelnet,centerpoint,bevfusion_icra23}, we collapse the height of the fused voxel features $F_f$ to transform them into BEV space. Then, we follow \cite{bevfusion_icra23} to introduce a convolution-based BEV encoder containing several convolutional layers to compensate for spatial misalignments between the LiDAR and camera caused by inaccurate depth during view transformation.

Following prior works \cite{bevfusion_icra23,bevfusion_nips22,transfusion,msmdfusion}, we use the same detection head as in \cite{transfusion} that is built upon transformer architecture \cite{attention}, which is first proposed in DETR \cite{detr}. The detection head consists of a transformer decoder and a feed-forward network. A set of embeddings (called object queries) \cite{attention} serves as the input for the decoder. During the decoder, the object queries are passed through a self attention module, then cross attention is conducted with our final fused BEV features $F_B$. The cross attention between object queries and the feature maps aggregates relevant context onto the object candidates, while the self attention between object queries reasons pairwise relations between different object candidates. After the decoder, the object queries containing rich instance information are then independently decoded into boxes and class labels by a feed-forward network (FFN).

Following UVTR \cite{uvtr}, the Hungarian algorithm is applied to match the predictions and the ground truth during training. Meanwhile, Focal loss \cite{focalloss} and \(L1\) loss are respectively used for the classification and 3D bounding box regression, the parameters of all mentioned modules are optimized together with the detector.

\vspace{-3pt}
\subsection{Pre-Fusion} 

\label{prefusion}
The Pre-Fusion consists of two key modules: the VEM and the IEM. These two modules work together to enhance the semantic awareness of LiDAR features and the 3D spatial awareness of camera features, characterized by bidirectional complementarity, allowing
for a more comprehensive 3D representation.

\begin{figure}[t!]
      \centering
      \includegraphics[width=\columnwidth]{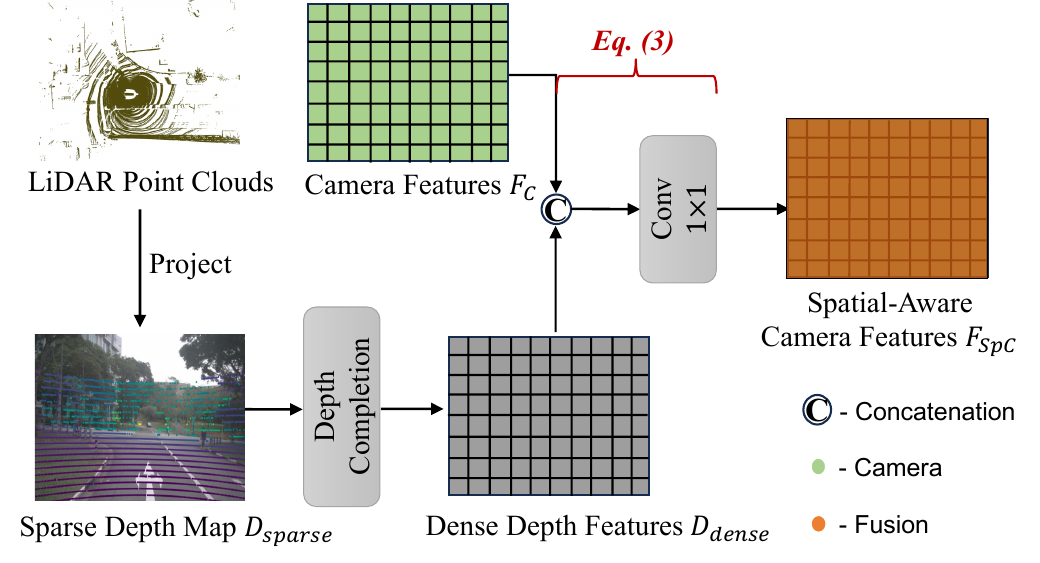}
      \caption{\textbf{Image Enhancement Module (IEM).} It first projects raw LiDAR points to the image coordinates. Then, we use depth completion to generate a dense depth feature map, whose size is the same as the camera
      features. Finally, the dense depth feature map is concatenated with the camera features, followed by a convolutional layer to reduce the feature dimension.}
      \label{iem} 
      \vspace{-16pt}
\end{figure}

\subsubsection{Voxel Enhancement Module}
\label{vem_details}
Despite having strong spatial localization capabilities, the LiDAR voxel features \(F_{L}\) obtained from the LiDAR branch perform poorly on detailed semantic information. This drawback can propagate through the Unified Fusion process. To overcome this issue and provide more comprehensive features for the Unified Fusion, we design the VEM, as shown in Fig. \ref{vem}.

In the VEM, the center point of every non-empty voxel is first projected onto image coordinates using the intrinsic and extrinsic parameters. To mitigate incorrect alignments, we then retrieve the \(K\) nearest camera features \(F_{nearest} \in \mathbb{R}^{K \times C_{2D}}\) around the projection point. During fusion, the contribution of these camera features should be different, so we designed a distance-prior weighting scheme. Specifically, for the \(K\) camera features, we first calculate the reciprocals of the distances to the projection point, denoted as \(L_{nearest} \in \mathbb{R}^{1 \times K}\), which is formed as the weights for \(F_{\text{nearest}}\). Then, we obtain the distance-prior weighted camera features \(F_{weighted} \in \mathbb{R}^{1 \times C_{2D}}\) as: 
\vspace{-1pt}
\begin{equation}
\vspace{-1pt}
F_{weighted} = \operatorname{Softmax}(L_{nearest}) \times F_{nearest} \text{,}
\end{equation}
where the Softmax function is used to calculate the weights of different camera features.

Finally, we use a Linear layer and an activation function followed by addition to achieve a learnable fusion process, resulting in the semantic-aware voxel features, denoted as \(F_{SeL}\). The whole process can be defined by:
\vspace{-1pt}
\begin{equation}
\vspace{-1pt}
F_{SeL} = \operatorname{ReLU}(\operatorname{Linear}(F_{weighted})) + F_{L} \text{.}
\end{equation}

\subsubsection{Image Enhancement Module} 
Camera features lack the ability to perceive accurate 3D spatial information, which is crucial for 3D object detection tasks. To enhance them from this view, we propose IEM, as shown in Fig. \ref{iem}.
We first project the point cloud onto the image plane to obtain a sparse depth map \(D_{sparse}\). Inspired by  \cite{deepinteraction}, we then use depth completion \cite{depth} followed by feature extraction to obtain a dense depth feature map \(D_{dense} \in \mathbb{R}^{H \times W \times C_{depth}}\). 
Finally, we concatenate the dense depth feature map with the camera features \(F_C \in \mathbb{R}^{H \times W \times C_{2D}}\) and use a convolutional layer to fuse them, getting the spatial-aware camera features \(F_{SpC}\):
\vspace{-1pt}
\begin{equation}
\vspace{-1pt}
F_{SpC} = \operatorname{Conv}(\operatorname{Concat}(F_C,D_{dense})) \text{.}
\end{equation}

Due to the geometric capability of the spatial-aware camera features, lifting them into 3D space ensures a more accurate process, providing a solid foundation for the subsequent Unified Fusion.

\subsection{Unified Fusion}
\label{unifiedfusion}
After obtaining the semantic-aware LiDAR voxel features and spatial-aware camera features, which have undergone specific bidirectional complementarity during Pre-Fusion, we further fuse them in the U-Fusion stage to build a semantic- and spatial-aware unified representation.

Previous work \cite{bevfusion_icra23,bevfusion_nips22} unifies the two modalities in the BEV space by view transformation. While this improved efficiency, it has an inherent limitation that prevents the fine-grained fusion which is vital for better performance. Therefore, after elevating the camera features to the voxel representation, we preserve its state, obtaining $\hat{F}_{SpC}$.

Given the enhanced features from both LiDAR (\(F_{SeL} \in \mathbb{R}^{X \times Y \times Z \times C_{3D}}\)) and Camera (\(\hat{F}_{SpC} \in \mathbb{R}^{X \times Y \times Z \times C_{2D}}\)) with same spatial dimensions, a straightforward idea is to directly concatenate them. Inspired by \cite{openoccupancy}, to dynamically select features from both modalities, we have used an adaptive weighting method to perform Unified Fusion as:
\vspace{-2pt}
\begin{align}
\alpha &= \mathcal{C}_{3D}
(\operatorname{Concat}(\mathcal{C}_{3D}(F_{SeL}),\mathcal{C}_{3D}(\hat{F}_{SpC}))), \\
F_{f} &= \sigma(\alpha) \cdot F_{SeL} + (1 - \sigma(\alpha)) \cdot \hat{F}_{SpC},
\end{align}
where \(\mathcal{C}_{3D}\) refers to 3D convolution, \(\sigma\) denotes \textit{Sigmoid} activation, and \(F_f\) is the fused unified features in voxel space.

\section{EXPERIMENTS}

\begin{table*}[h]
\centering
\captionsetup{justification=centering, font=small}
\caption{\textbf{Comparison with SOTA methods on nuScenes test set.} ‘L’ and ‘C’ denote the LiDAR and the camera. ‘C.V.’, ‘Motor.’, ‘Ped.’, and ‘T.C.’ refer to the construction vehicle, motorcycle, pedestrian, and traffic cone, respectively. ‘\dag’ denotes the model with test-time augmentation and model ensemble.}
\vspace{-6pt}
\label{sota}
\begin{tabular}{c|c|cc|cccccccccc}
\toprule
Method & Modality & mAP & NDS & Car & Truck & C.V. & Bus & Trailer & Barrier & Motor. & Bike & Ped. & T.C. \\ 
\midrule
PointPillars \cite{pointpillars}& L & 40.1 & 55.0 & 76.0 & 31.0 & 11.3 & 32.1 & 36.6 & 56.4 & 34.2 & 14.0 & 64.0 & 45.6 \\ 

CenterPoint \cite{centerpoint}& L & 60.3 & 67.3 & 85.2 & 53.5 & 20.0 & 63.6 & 56.0 & 71.1 & 59.5 & 30.7 & 84.6 & 78.4 \\ 

TransFusion-L \cite{transfusion}& L & 65.5 & 70.2 & 86.2 & 56.7 & 28.2 & 66.3 & 58.8 & 78.2 & 68.3 & 44.2 & 86.1 & 82.0 \\ 

FocalFormer3D \cite{focalformer3d}& L & 68.7 & 72.6 & 87.2 & 57.1 & 34.4 & 69.6 & 64.9 & 77.8 & 76.2 & 49.6 & 88.2 & 82.3 \\ 
\midrule
Pointpainting \cite{pointpainting}& L+C & 46.4 & 58.1 & 77.9 & 35.8 & 15.8 & 36.2 & 37.3 & 60.2 & 41.5 & 24.1 & 73.3 & 62.4 \\ 

MVP \cite{mvp}& L+C & 66.4 & 70.5 & 86.8 & 58.5 & 26.1 & 67.4 & 57.3 & 74.8 & 70.0 & 49.3 & 89.1 & 85.0 \\ 

PointAugmenting \cite{pointaugmenting}& L+C & 66.8 & 71.0 & 87.5 & 57.3 & 28.0 & 65.2 & 60.7 & 72.6 & 74.3 & 50.9 & 87.9 & 83.6 \\ 

UVTR \cite{uvtr}& L+C & 67.1 & 71.1 & 87.5 & 56.0 & 33.8 & 67.5 & 59.5 & 73.0 & 73.4 & 54.8 & 86.3 & 79.6 \\ 

AutoAlignV2 \cite{autoalignv2}& L+C & 68.4 & 72.4 & 87.0 & 59.0 & 33.1 & 69.3 & 59.3 & - & 72.9 & 52.1 & 87.6 & - \\ 

TransFusion-LC \cite{transfusion}& L+C & 68.9 & 71.7 & 87.1 & 60.0 & 33.1 & 68.3 & 60.8 & 78.1 & 73.6 & 52.9 & 88.4 & 86.7 \\ 

BEVFusion \cite{bevfusion_nips22}& L+C & 69.2 & 71.8 & 88.1 & 60.9 & 34.4 & 69.3 & 62.1 & 78.2 & 72.2 & 52.2 & 89.2 & 85.2 \\ 

BEVFusion \cite{bevfusion_icra23}& L+C & 70.2 & 72.9 & \underline{88.6} & 60.1 & \underline{39.3} & 69.8 & 63.8 & 80.0 & 74.1 & 51.0 & 89.2 & 86.5 \\ 

DeepInteraction \cite{deepinteraction}& L+C & 70.8 & 73.4 & 87.9 & 60.2 & 37.5 & 70.8 & 63.8 & 80.4 & 75.4 & 54.5 & 90.3 & 87.0 \\ 

UniTR \cite{unitr}& L+C & 70.9 & 74.5 & 87.9 & 60.2 & 39.2 & 72.2 & 65.1 & 76.8 & 75.8 & 52.2 & 89.4 & \underline{89.7} \\ 

MSMDFusion \cite{msmdfusion}& L+C & 71.5 & 74.0 & 88.4 & 61.0 & 35.2 & 71.4 & 64.2 & \underline{80.7} & 76.9 & 58.3 & 90.6 & 88.1 \\ 

FocalFormer3D \cite{focalformer3d}& L+C & 71.6 & 73.9 & 88.5 & 61.4 & 35.9 & 71.7 & \underline{66.4} & 79.3 & \underline{80.3} & 57.1 & 89.7 & 85.3 \\ 

SparseFusion \cite{sparsefusion}& L+C & 72.0 & 73.8 & 88.0 & 60.2 & 38.7 & 72.0 & 64.9 & 79.2 & 78.5 & 59.8 & \underline{90.9} & 87.9 \\ 

CMT \cite{cmt}& L+C & 72.0 & 74.1 & 88.0 & \underline{63.3} & 37.3 & \underline{75.4} & 65.4 & 78.2 & 79.1 & \underline{60.6} & 87.9 & 84.7 \\ 
\midrule
BiCo-Fusion~(Ours) & L+C & \underline{72.4} & \underline{74.5} & 88.1 & 61.9 & 38.2 & 73.3 & 65.7 & 80.4 & 78.9 & 59.8 & 89.7 & 88.3 \\ 

BiCo-Fusion\dag~(Ours) & L+C & \textbf{76.1} & \textbf{77.1} & \textbf{89.6 }& \textbf{67.4} & \textbf{44.1} & \textbf{76.4} & \textbf{66.8} & \textbf{83.5} & \textbf{82.6} & \textbf{65.9} & \textbf{93.6} & \textbf{90.8} \\
\bottomrule
\end{tabular}
\vspace{-12pt}
\end{table*}

\begin{figure*}

    \centering  
    \includegraphics[width=.97\linewidth]{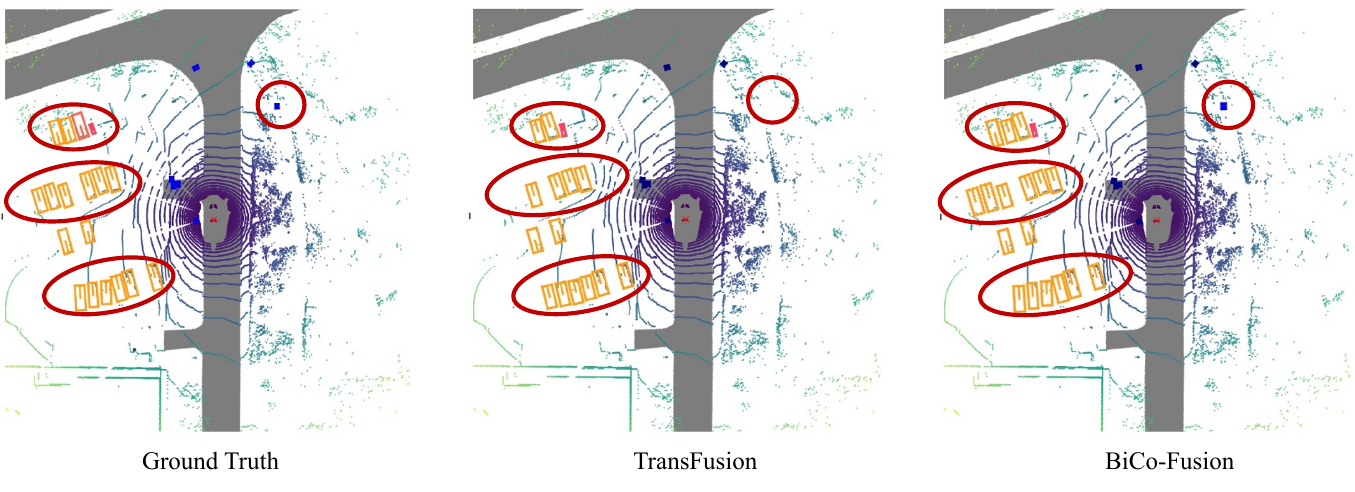}
     \vspace{-10pt}
     \caption{\textbf{Comparison of the qualitative results between our method and others on nuScenes validation set.} We provide results of ground truth, TransFusion, and our BiCo-Fusion. Results on cars are colored in yellow, pedestrians in blue, and trucks in red. The red circles highlight the difference among these results. We can observe BiCo-Fusion can detect more occluded objects and some small objects, which indicates the superiority of our method.}
    \label{compare}
    \vspace{-14pt}
\end{figure*}

\subsection{Experimental Setup}

\noindent \textbf{Dataset.} Following\cite{bevfusion_icra23,transfusion}, we evaluate the 3D object detection performance of our BiCo-Fusion on the nuScenes benchmark \cite{nuscenes}. The nuScenes dataset is a large-scale autonomous driving benchmark including 1,000 driving scenarios in total (700, 150, and 150 scenes for training, validation, and testing, respectively). To evaluate 3D object detection, there are two most important metrics: mean Average Precision (mAP) and nuScenes detection scores (NDS), which reflect the total performance of a method. The mAP is calculated from the mean of the average precision (AP) across ten classes under distance thresholds of 0.5m, 1m, 2m, and 4m. NDS is a robust metric that integrates mAP with five True Positive (TP) metrics: mATE, mASE, mAOE, mAVE, and mAAE, which together ensure a comprehensive evaluation of object translation, scale, orientation, velocity, and attributes.

\begin{table}[t!]
    \centering
    \captionsetup{justification=centering, font=small}
    \caption{Comparison on the nuScenes validation set.}
     \vspace{-6pt}
     \setlength{\tabcolsep}{5.8mm}{
    \begin{tabular}{c|cc|c}
    \toprule
      Method  &  mAP  & NDS & FPS\\
       \midrule
       BEVFusion \cite{bevfusion_nips22} & 67.9 & 71.0 & 0.8 \\
       TransFusion \cite{transfusion}  & 67.5 & 71.3 & 3.4 \\
       BEVFusion \cite{bevfusion_icra23} & 68.5 & 71.4 & \textbf{4.3} \\
       \midrule
       BiCo-Fusion (Ours) & \textbf{70.5} & \textbf{72.9} & 2.8 \\ 
      \bottomrule 
    \end{tabular}
    }
    \label{fps}
    \vspace{-15pt}
\end{table}

\noindent \textbf{Implementation Details.} Our implementation is based on the MMDetection3D framework \cite{mmdet3d2020}. We use Swin-T \cite{swintransformer} as our image backbone and VoxelNet \cite{voxelnet} as our LiDAR backbone. We set the image size to 384\(\times\)1056 and voxelize the LiDAR point cloud with 0.075m, and the point cloud covers [-54m, 54m] along the X and Y axes, [-5m, 3m] along the Z axis, respectively. Our model is trained with a batch size of 16. Following \cite{transfusion}, our training consists of two stages: (1) We first train the LiDAR-only baseline for 20 epochs. (2) We then finetune the proposed LiDAR-camera fusion framework for another 6 epochs. We follow CBGS \cite{cbgs} to perform class-balanced sampling and employ the AdamW optimizer \cite{adamw} with one cycle learning rate policy \cite{onecycle}, with max learning rate \(1e^{-3}\), weight decay 0.01. 

\begin{table}[t!]
\centering
\captionsetup{justification=centering, font=small}
\caption{Ablation studies for our proposed components on the nuScenes validation set.}
\vspace{-6pt}
\label{component-wise}
\setlength{\tabcolsep}{2.5mm}{
\begin{tabularx}{\columnwidth}{c|cccc|cc}
\toprule
 & Baseline & VEM & IEM & U-Fusion & mAP & NDS \\ 
\midrule
(1) & \checkmark   &  &   &  & 64.6 & 69.3 \\
(2) & \checkmark & \checkmark  &   &   &  66.5  & 70.3 \\
(3) & \checkmark &  & \checkmark & \checkmark & 69.2 & 72.0 \\
(4) & \checkmark & \checkmark &   & \checkmark & 70.0 & 72.5 \\
(5) & \checkmark & \checkmark & \checkmark  & \checkmark & \textbf{70.5} & \textbf{72.9} \\
\bottomrule
\end{tabularx}}
\vspace{-18pt}
\end{table}

We additionally combine BiCo-Fusion with test-time augmentation (TTA) and model ensemble techniques \cite{centerpoint} to get the BiCo-Fusion\dag. Specifically, our BiCo-Fusion\dag~ensembles several BiCo-Fusion models with different voxel sizes ranging from (0.05m, 0.05m, 0.2m) to (0.125m, 0.125m, 0.2m), with intervals of 0.025m. Each model is adopted with standard test-time augmentation like double flipping and rotations (\ie, {0°, ±22.5°, ±180°}) to the input point clouds. About the details on how to combine TTA and model ensemble to produce the final detection, our implementation is consistent with prior works~\cite{centerpoint,focalformer3d,bevfusion_icra23,cmt}. Specifically, we first train these multiple models with different voxel sizes. During inference, TTA is applied to each model, where the input data undergoes the above transformations to generate multiple augmented samples. Then each model produces predictions for these augmented samples. Following~\cite{cmt,centerpoint}, we use weighted box fusion (WBF) \cite{wbf} to fuse these predictions, obtaining a single prediction per model. Finally, for model ensemble, we use non-maximum suppression (NMS) \cite{fasterrcnn} to ensemble predictions from all models to produce the final result.

\subsection{Main Results}

In the fair setting without using TTA and model ensemble, we compare our BiCo-Fusion with leading methods on nuScenes test set, as shown in Tab.~\ref{sota}. The class-specific APs of BiCo-Fusion stay very close to some very recent works, including \cite{cmt} and \cite{sparsefusion}. More importantly, BiCo-Fusion improves the LiDAR-only baseline \cite{transfusion} by 6.9\% mAP and 4.3\% NDS with the additional fusion of multi-modalities, achieving superior performance with 72.4\% mAP and 74.5\% NDS, which demonstrates we outperform all existing SOTA 3D detection methods with better average performance across all classes, and better overall performance under comprehensive evaluation setting of the NDS \cite{nuscenes}. As an example, although CMT~\cite{cmt} gets the highest APs in some classes (Truck, Bus, Bike), our method outperforms it in more classes (Car, C.V., Trailer, Barrier, Ped., T.C.), which makes our method superior in average performance. By simply adding TTA and model ensemble techniques, which are widely adopted by prior works for submissions on the nuScenes leaderboard \cite{centerpoint,bevfusion_icra23,deepinteraction,focalformer3d,msmdfusion,cmt}, the BiCo-Fusion\dag~version achieves a huge performance improvement with 76.1\% mAP and 77.1\% NDS, reflecting the scalability of our method.

\begin{table}[t!]
    \centering
    \captionsetup{justification=centering, font=small}
    \caption{Parameter selection in Voxel Enhancement Module.}
    \vspace{-6pt}
    \setlength{\tabcolsep}{5.3mm}{
    \begin{tabular}{c|cc}
    \toprule
      Number of camera surroundings (K)  &  mAP  & NDS \\
       \midrule
       1  &  69.2 & 71.9 \\
       3  &  69.8 & 72.4 \\
       6  &  70.3 & 72.7 \\
       9  &  \textbf{70.5} & \textbf{72.9} \\
       10 &  70.4 & 72.9 \\
      \bottomrule 
    \end{tabular}}
    \label{k in vem}
    \vspace{-10pt}
\end{table}
\begin{table}[t!]
    \centering
    \captionsetup{justification=centering, font=small}
    \caption{Effect of Distance-prior weighting scheme in VEM.}
     \vspace{-6pt}
     \setlength{\tabcolsep}{6.6mm}{
    \begin{tabular}{c|cc}
    \toprule
       Weighting scheme in VEM &  mAP  & NDS \\
       \midrule
       w/o distance-prior  &  69.7 & 72.3 \\
       w/ distance-prior  &  \textbf{70.5} & \textbf{72.9} \\
      \bottomrule 
    \end{tabular}}
    \label{dp in vem}
    \vspace{-8pt}
\end{table}
As shown in Tab.~\ref{fps}, BiCo-Fusion achieves superior 3D detection performance on the nuScenes validation set while maintaining a comparable inference speed. Fig. \ref{compare} presents the qualitative results on the nuScenes validation set. Our method predicts more accurate results in some complex cases like the objects are occluded or many objects are tightly packed together.

\subsection{Ablation Studies}
\subsubsection{Component-wise Ablation}
As shown in Tab. \ref{component-wise}, we conduct ablation studies on our proposed modules on the nuScenes validation set. In Tab. \ref{component-wise} (1), we reimplement TransFusion-L \cite{transfusion} as our LiDAR-only baseline. According to Tab. \ref{component-wise} (2), after combining VEM, we improve mAP by 1.9\% and NDS by 1\%, which indicates the effectiveness of enhancing the semantic capability of voxel features. In Tab.~\ref{component-wise} (3), combining IEM and U-Fusion, the mAP and NDS improve to 69.2\% and 72.0\%, respectively, which brings a large shift. But the overall performance is lower than the combination of VEM and U-Fusion, suggesting that enhancing the semantic awareness of the LiDAR modality (\ie, VEM) leads to better improvement than enhancing the 3D spatial awareness of the camera modality (\ie, IEM). Incorporating the Unified Fusion (U-Fusion) with VEM (Tab.~\ref{component-wise} (4)) leads to additional improvement by 3.5\% mAP and 2.2\% NDS compared to using only VEM. It should be emphasized that there are two main reasons why the combination contributes to such a big improvement: a) The Voxel Enhancement Module in Pre-Fusion prepares more comprehensive LiDAR features, setting the stage for the second step of fusion. b) The fusion in a unified voxel space fills in many empty voxels, compensating for the geometric sparsity of the LiDAR modality. 

Finally, with all these components combined (Tab. \ref{component-wise} (5)), the final mAP and NDS achieve significant enhancement to 70.5\% and 72.9\%, respectively, demonstrating the effectiveness of our whole framework, which ensures semantic- and spatial-aware 3D object detection.

\subsubsection{Discussion on Voxel Enhancement Module} 
We analyze the selection of the number of surrounding camera features per voxel (\(K\)) in the voxel enhancement module. As shown in Tab. \ref{k in vem}, when \(K\) increases from 1 to 9, the mAP and NDS gradually rise. And when \(K\) is increased to 10, there is no change in NDS, but the mAP decreases, so we select \(K = 9\) for the framework.

Tab. \ref{dp in vem} demonstrates the effectiveness of our distance-prior weighting scheme in the voxel enhancement module, which improves the mAP and NDS by 0.8\% and 0.6\%, respectively.

\subsubsection{Discussion on Unified Fusion} In Unified Fusion (U-Fusion), we employ an adaptive weighting scheme, which dynamically fuses features from both modalities to build a unified representation. We ablate it in Tab. \ref{aw in uf}, which indicates that our scheme improves performance compared with simple concatenation, by 1.1\% mAP and 0.7\% NDS.

\begin{table}[t!]
    \centering
    \captionsetup{justification=centering, font=small}
    \caption{Effect of adaptive weighting in U-Fusion.}
    \vspace{-6pt}
    \setlength{\tabcolsep}{5.9mm}{
    \begin{tabular}{c|cc}
    \toprule
       Weighting scheme in U-Fusion &  mAP  & NDS \\
       \midrule
       w/o adaptive weighting  &  69.4 & 72.2 \\
       w/ adaptive weighting  &  \textbf{70.5} & \textbf{72.9} \\
      \bottomrule 
    \end{tabular}}
    \label{aw in uf}
    \vspace{-8pt}
\end{table}

\begin{table}[t!]
\vspace{-1pt}
\centering
\captionsetup{justification=centering, font=small}
\caption{Ablation results under different settings.}
\vspace{-6pt}
\setlength{\tabcolsep}{3.5mm}{
\begin{tabular}{l|c|cc}
\toprule
  & Different Choices & mAP & NDS \\
\midrule
\multirow{2}{*}{(a) Image backbone} & ResNet-50 & 69.4 & 72.1 \\
 & ResNet-101 & 70.1 & 72.6 \\
 & Swin-T & \textbf{70.5} & \textbf{72.9} \\
\midrule
\multirow{2}{*}{(b) Voxel size} & 0.075 & \textbf{70.5} & \textbf{72.9} \\
 & 0.1 & 70.0 & 72.5 \\
\midrule
\multirow{2}{*}{(c) Image size} & 256 x 704 & 69.9 & 72.5 \\
 & 384 x 1056 & \textbf{70.5} & \textbf{72.9} \\
\bottomrule
\end{tabular}}
\label{frameworkchoices}
\vspace{-16pt}
\end{table}

\subsubsection{Ablation on Different Choices in Framework} We further construct ablation studies on different choices in our BiCo-Fusion framework. As shown in Tab. \ref{frameworkchoices} (a), our framework can benefit from more powerful image backbones. Swin-T \cite{swintransformer} achieves the best performance, with improving mAP by 1.1\%, compared with ResNet \cite{resnet}. Tab. \ref{frameworkchoices} (b) demonstrates that a smaller voxel size can bring a small-scale improvement, with about 0.5\% for both mAP and NDS.
Same as voxel size, using a larger image resolution (\textit{e.g.}, 384 × 1056) also leads to a slight performance improvement, as shown in Tab. \ref{frameworkchoices} (c).

\vspace{-4pt}
\section{CONCLUSION}
This letter presented BiCo-Fusion, a novel LiDAR-camera fusion framework for achieving semantic- and spatial-aware 3D object detection.
In BiCo-Fusion, we introduced Pre-Fusion with a bidirectional complementary strategy to enhance the semantic awareness of LiDAR features and the 3D spatial awareness of camera features, which ensures comprehensiveness at the semantic and spatial levels for both modalities, and also Unified Fusion (U-Fusion) to adaptively fuse the enhanced features from both modalities for building a unified representation. 
Extensive experiments demonstrated the effectiveness of these components and our proposed framework.

\noindent \textbf{Limitations and Future Work.} At present, BiCo-Fusion uses a simple projection and retrieval mechanism in Pre-Fusion, which relies on alignments on the nuScenes dataset. We believe that this can be better optimized by introducing a softer attention mechanism to avoid some calibration inaccuracies. Besides, a supervised depth estimation \cite{bevdepth} is able to further enhance the spatial structure of the final 3D representation from the camera branch, which ensures better performance. We leave these limitations as future work.




\bibliographystyle{IEEEtranBST/IEEEtran}
\bibliography{IEEEtranBST/IEEEabrv, IEEEtranBST/reference}

\end{document}